\pdfoutput=1
\documentclass[sigconf]{acmart}

\usepackage{subfigure}
\usepackage{hyperref}
\usepackage{array}

\AtBeginDocument{%
  \providecommand\BibTeX{{%
    \normalfont B\kern-0.5em{\scshape i\kern-0.25em b}\kern-0.8em\TeX}}}

\setcopyright{rightsretained}
\copyrightyear{2019}
\acmYear{2019}
\acmDOI{10.1145/nnnnnnn.nnnnnnn}

\acmConference[KDD Workshop on AI for fashion]{Anchorage '19:  KDD Workshop on AI for fashion}{August 05, 2019}{Anchorage, Alaska}
\acmBooktitle{KDD Workshop on AI for fashion,
  August 05, 2019,  Anchorage, Alaska}
\acmPrice{}
\acmISBN{978-x-xxxx-xxxx-x/YY/MM}

\begin{document}
\settopmatter{printfolios=True}
\title[Deep Neural Semantic Hashing]{Adversarially Trained Deep Neural Semantic Hashing Scheme for Subjective Search in Fashion Inventory}

\author{Saket Singh}
\email{saketsingh@iitkgp.ac.in}
\author{Debdoot Sheet}
\email{debdoot@ee.iitkgp.ac.in}
\affiliation{%
  \institution{Indian Institute of Technology Kharagpur}
  \city{Kharagpur}
  \state{West Bengal}
  \country{India}
  \postcode{721302}
}

\author{Mithun Dasgupta}
\affiliation{%
  \institution{Microsoft}
  \city{Hyderabad}
  \country{India}}
\email{migupta@microsoft.com}


\begin{abstract}
  The simple approach of retrieving a closest match of a query image from one in the gallery, compares an image pair using sum of absolute difference in pixel or feature space. The process is computationally expensive, ill-posed to illumination, background composition, pose variation, as well as inefficient to be deployed on gallery sets with more than 1000 elements. Hashing is a faster alternative which involves representing images in reduced dimensional simple feature spaces. Encoding images into binary hash codes enables similarity comparison in an image-pair using the Hamming distance measure. The challenge however lies in encoding the images using a semantic hashing scheme that lets subjective neighbours lie within the tolerable Hamming radius. This work presents a solution employing adversarial learning of a deep neural semantic hashing network for fashion inventory retrieval. It consists of a feature extracting convolutional neural network (CNN) learned to (i) minimize error in classifying type of clothing, (ii) minimize hamming distance between semantic neighbours and maximize distance between semantically dissimilar images, (iii) maximally scramble a discriminator's ability to identify the corresponding hash code-image pair when processing a semantically similar query-gallery image pair. Experimental validation for fashion inventory search yields a mean average precision (mAP) of 90.65\% in finding the closest match as compared to 53.26\% obtained by the prior art of deep Cauchy hashing for hamming space retrieval.
\end{abstract}

%
%






\maketitle

\section{Introduction}
\begin{figure}[t]
\begin{center}
\includegraphics[width=1.\linewidth]{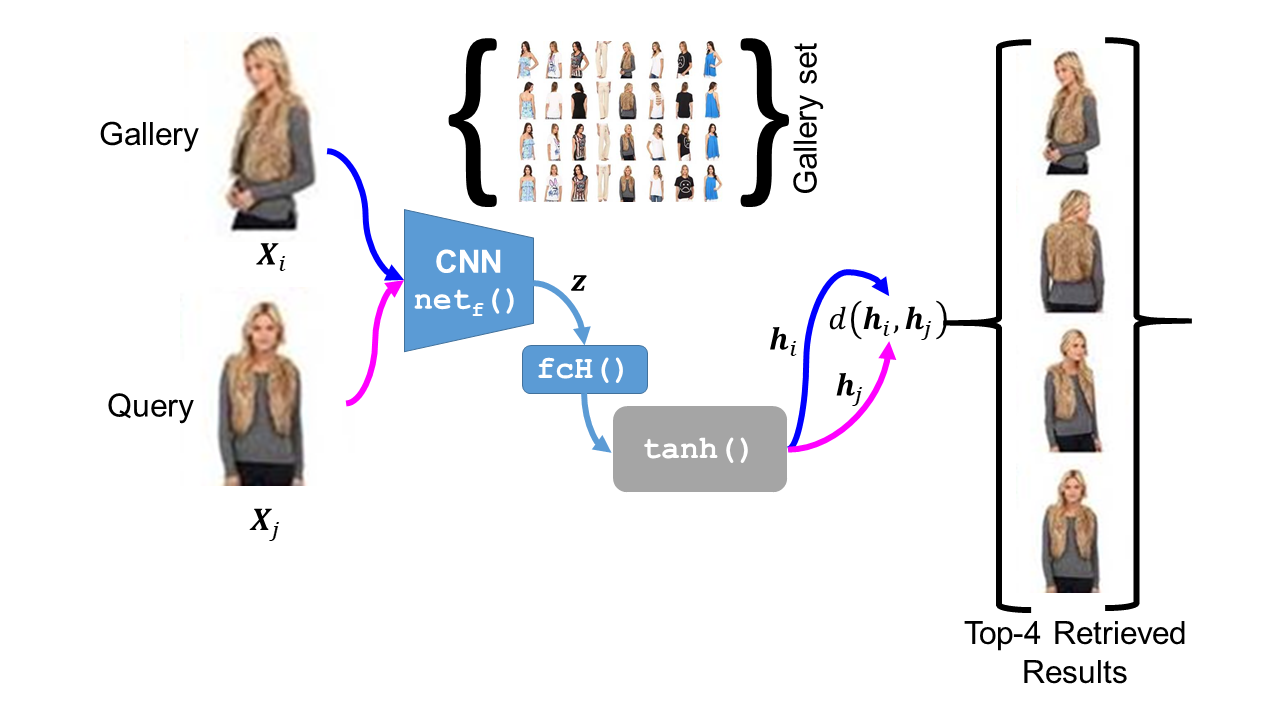}
\end{center}
   \caption{Approach of semantic hashing based retrieval, where the CNN generates continuous value feature vector $\mathbf{z}$ corresponding to the query $\mathbf{X}_j$ which is subsequently binarized to yield the hash code $\mathbf{h}_j$. Hamming distance $d(\mathbf{h}_i,\mathbf{h}_j)$ is computed to measure similarity with an image $\mathbf{X}_i$ from the gallery set. The process yields subjectively similar results from the gallery independent of pose and other invariances.}
\label{fig:hashing}
\end{figure}

Retrieval of subjectively similar results such as in images becomes challenging in the era of big data, when its computationally challenging to employ pixel-wise or feature-wise image-paid difference measures for comparison in very large datasets. The concept of semantic hashing~\cite{salakhutdinov2009semantic} was introduced in this regards to be able to develop subjectively similar search for retrieval. With growth of e-commerce this has gained center stage with demand for retrieving subjectively similar fashion inventory. The concept is to be able to represent an image in terms of binary hash codes to be able to compute inexpensive similarity measures for fast pair-wise matching for retrieval. The caveat though is to be able to design hash codes where subjectively similar entries are within a tolerable radius of each other as expected in Fig.~\ref{fig:hashing}.

Hashing was originally used in cryptography to encode high dimensional data into smaller compact codes, sequences or strings using a derived hashing function. In image retrieval hashing involves encoding images  into a fixed length vector representation. The code vector is typically binary represented to enable use of the computationally inexpensive normalized Hamming distance for fast pair-wise comparison between a query with an image from the gallery. The challenge however being to achieve codes which enable subjectively similar images to be within a tolerable search neighbourhood, Cauchy probability function had been presented in earlier works\cite{Cao_2018_CVPR}. Similarity learning in simple words aims at generating similar hash codes for similar data whereas dissimilar data should show considerable variation in their hash codes~\cite{DBLP:journals/corr/LiWK15}. Here \emph{similar} may refer to visually similar or semantically or subjectively similar. Inspired by the robustness of convolutional neural networks (CNN)~\cite{LeCun:1989:BAH:1351079.1351090} in solving several computer vision tasks, in this paper we propose to train a CNN framework to produce binary hash codes under certain constraints defining its Hamming distance neighbourhood and pair wise relationship, expected during the comparison. 

The paper is organized to detail related prior work in Sec.~\ref{sec:relatedwork}, our proposed method in Sec.~\ref{sec:method}, experiments and results obtained thereof in Sec.~\ref{sec:expt}, followed by discussion of the results obtained in Sec.~\ref{sec:disc}, and conclusion of the work in Sec.~\ref{sec:conc}.

\section{Related Work}
\label{sec:relatedwork}

Supervised learning of CNNs for hashing of images have proven to be better at generating hash codes. They typically incorporate the class label information of an image to be able to learn features characteristic of each class of objects, viz. in case of search in fashion databases, different clothing types have characteristic features such as shirts have features characteristically different from trousers or skirts, etc. Recent works employ pair-wise image labels for generating effective hash functions. Such methods employing pair-wise similarity learning generally perform better~\cite{Cao_2018_CVPR,7780596} than non-similarity based hashing~\cite{7301269} which are easier while not requiring any label information for understanding similarity. 

Earlier approaches employing \emph{non-similarity matched hashing} employed image classification models such as with CNNs that were  modified to generate binary codes of features extracted in the penultimate layers, with use of functions like  \emph{sigmoid} or \emph{tanh} for generating binary codes from continuous valued data. The retrieval task typically is performed in two stages  as \emph{coarse} and \emph{fine}~\cite{7301269}. The coarse stage retrieves a large set of candidates using inexpensive distance measures like the Hamming distance. In the fine stage, the distance measures like Euclidean are employed on the continuous valued features for finding the closest match. 

Recent approaches in line with \emph{similarity matched hashing} have employed deep Cauchy hashing. This approach predicts the similarity label using \emph{Cauchy function} and also uses quantization loss to compensate the relaxation provided by the binary hash code generating function\cite{Cao_2018_CVPR}. Cauchy function has proved to be more effective than sigmoid in estimating optimal values of the similarity index and penalizing the losses obtained. Quantization loss ensures that the generated hash codes are close to exact limits of binary values\cite{DBLP:journals/corr/CaoL0Y17}, with the limitation being the large number of epochs required to train these networks.

Although, the supervised hashing methods, especially those employing deep learnt hash functions have showed remarkable performance in representing input data using binary codes, they require costly to acquire human-annotated labels for training. In absence of annotated large datasets, their performance significantly degrades. The unsupervised hashing methods on the other hand easily address this issue by providing learning frameworks that do not require any labelled input. Semantic hashing is one of the early studies, which adopts restricted Boltzmann machine (RBM) as a deep hash function~\cite{Hinton504}.

\section{Hashing Method for Subjectively Similar Search}
\label{sec:method}

\begin{figure}[t]
\begin{center}
\includegraphics[width=1.0\linewidth]{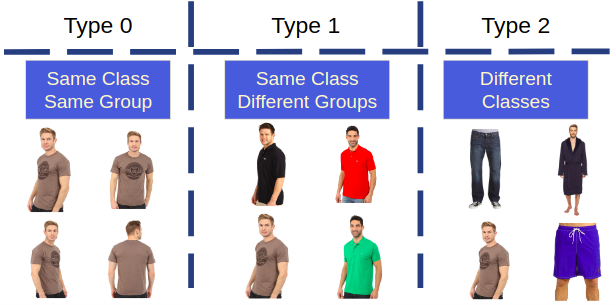}
\end{center}
   \caption{Figure shows the categorization of dataset.}
\label{fig:datacat}
\end{figure}

\begin{figure*}[t]
\begin{center}
    \begin{subfigure}[Overview of the learning scheme for semantic hashing.]{\includegraphics[width=0.48\textwidth]{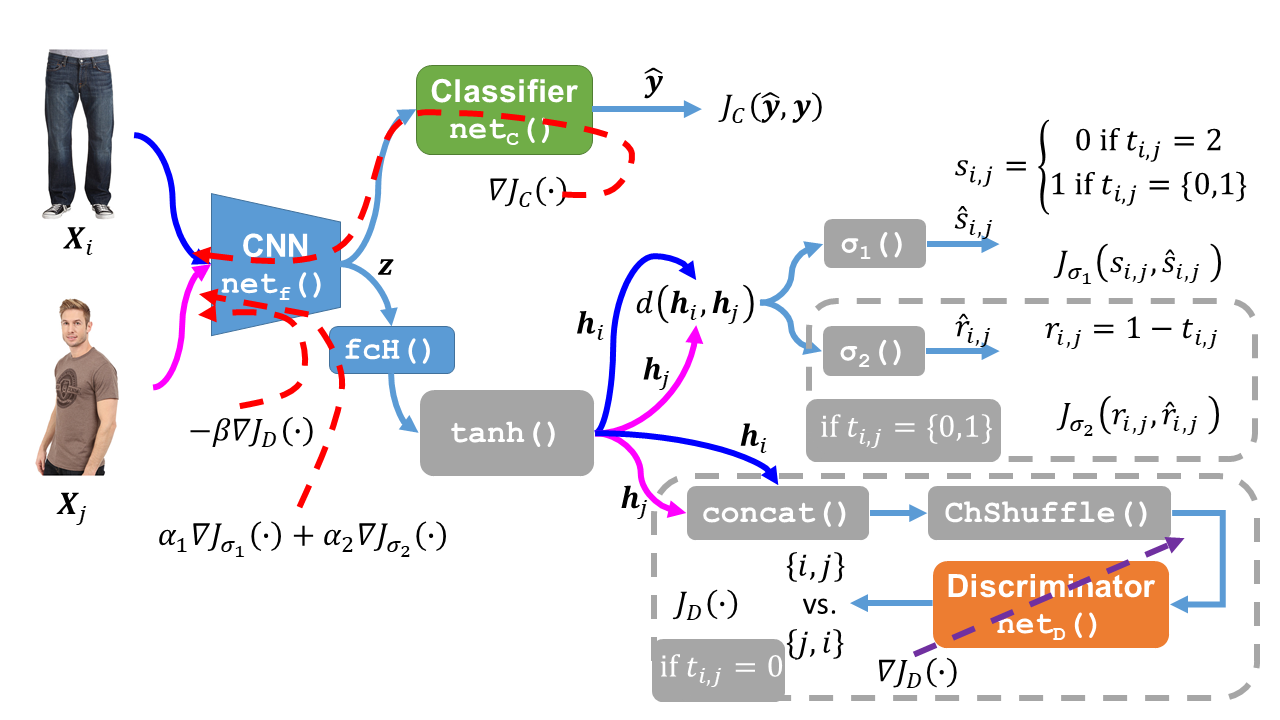}
        \label{fig3a}}
    \end{subfigure}
    \begin{subfigure}[Stage 1: Learning of clothing item features.]{\includegraphics[width=0.48\textwidth]{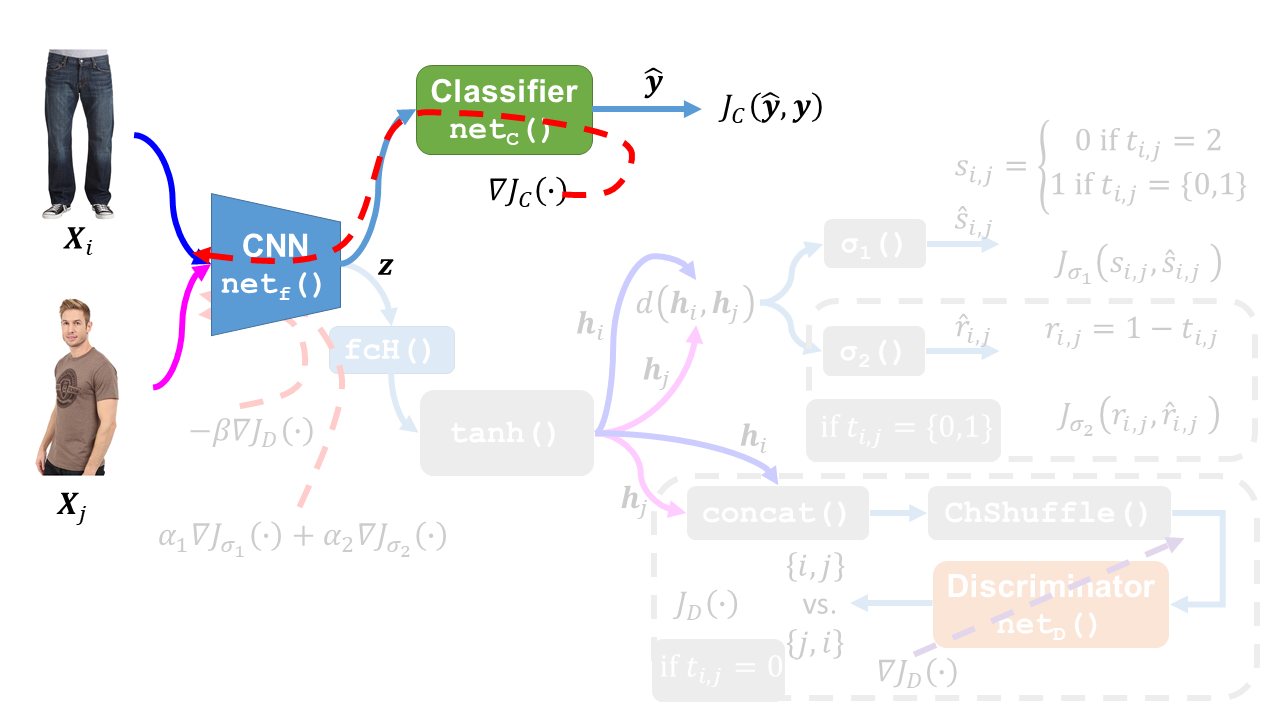}
        \label{fig3b}}
    \end{subfigure} 
    \begin{subfigure}[Stage 2: Learning with Cauchy similarity measure.]{\includegraphics[width=0.48\textwidth]{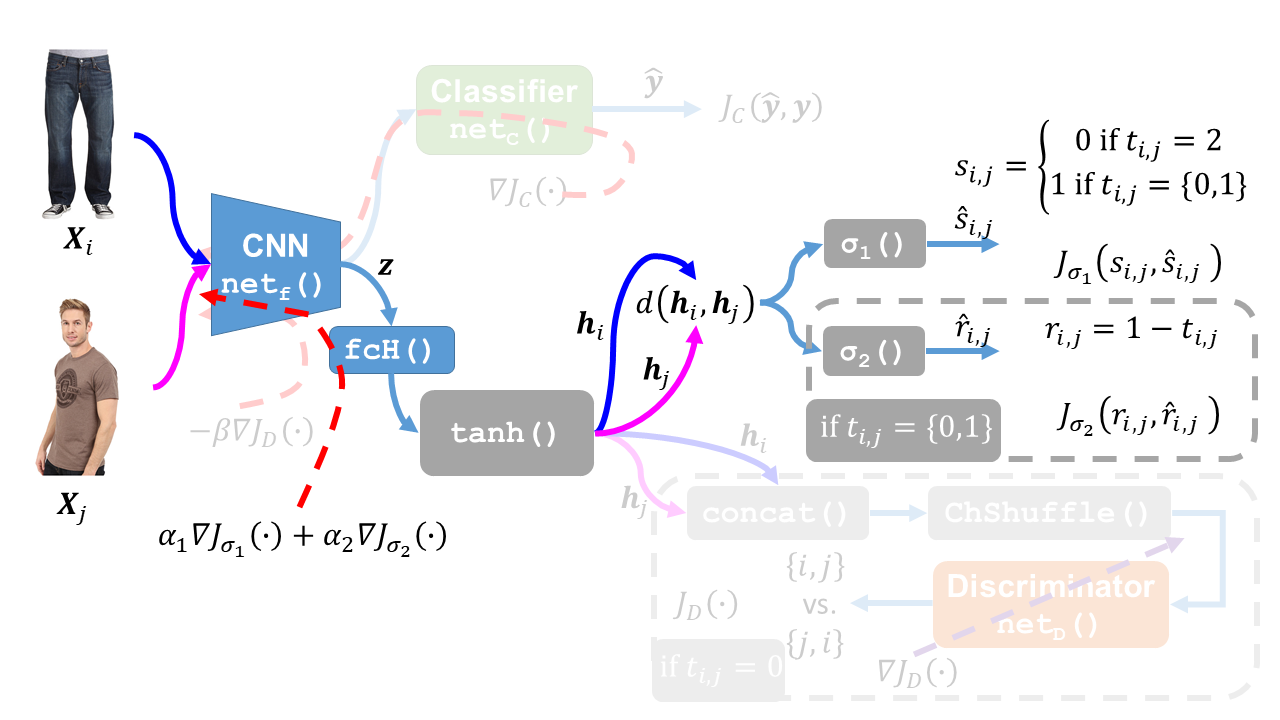}
        \label{fig3c}}
    \end{subfigure} 
    \begin{subfigure}[Stage 3: Adversarial learning of relational similarity.]{\includegraphics[width=0.48\textwidth]{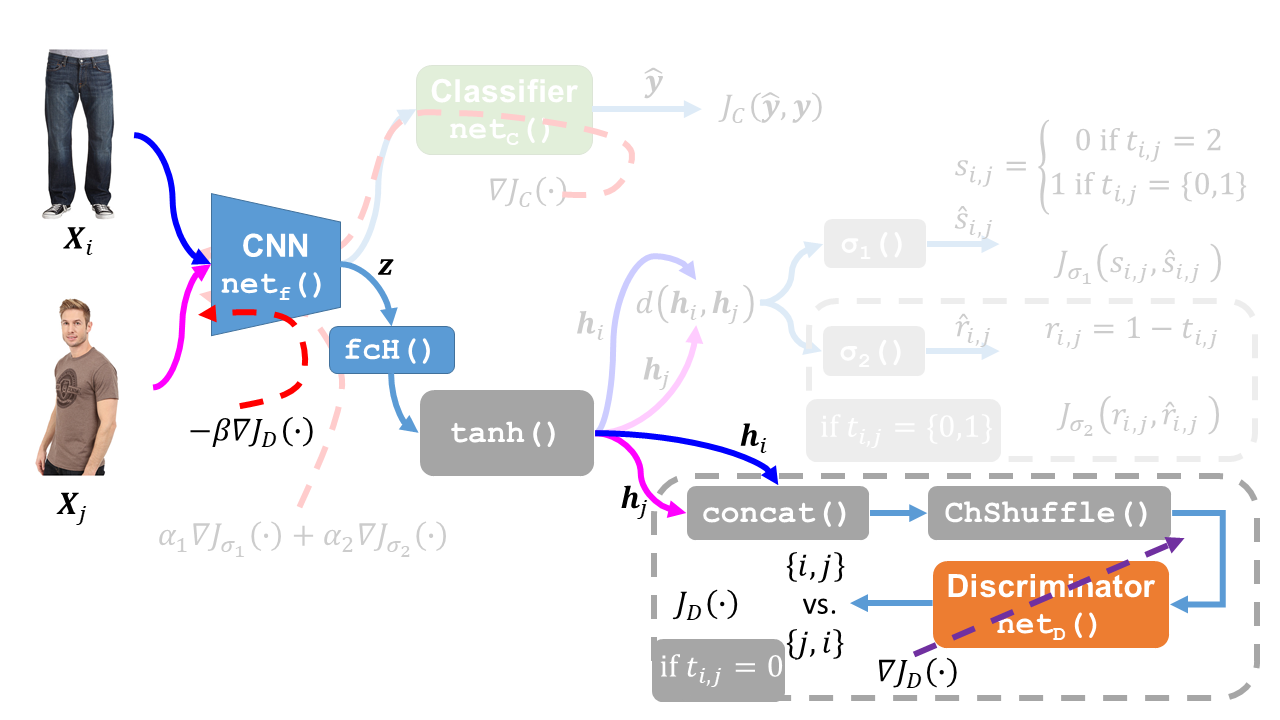}
        \label{fig3d}}
    \end{subfigure}
    \caption{Framework for learning of the deep neural semantic hashing scheme for subjective search across images. Blocks in gray represent units with non-learnable parameters.}
\label{fig:resultsnapshot}
\end{center}
\end{figure*}

In pairwise similarity based training the input is a pair of images along with their similarity index calculated based on their shared attributes that are obtained from their annotations. In this approach, the Cauchy probability function \cite{Cao_2018_CVPR} is used to predict the similarity label. Given an image $\mathbf{X}_i$ and another $\mathbf{X}_j$ in a pair, when they belong to the same class they are regarded as \emph{similar} and indicated with the similarity index $s_{i,j}=1$ and when they belong to different classes they are regarded as \emph{dissimilar} with $s_{i,j}=0$. As can be seen in Fig.~\ref{fig:datacat} this relationship is described in terms of view and pose variations across different types of clothes. \textbf{Type 0} indicates all images of the same item under different poses or background variations, and a pair selected from this set is represented as $t_{i,j}=0$. \textbf{Type 1} indicates same class of clothing item viz. only shirts but each of different color, and a pair selected from this set is represented as $t_{i,j}=1$. \textbf{Type 2} represents different classes of clothing items viz. shirts and shorts, etc. and a pair selected from this set is represented as $t_{i,j}=2$. Subjective similarity is defined as $s_{i,j}=0$ when $t_{i,j}=2$ and $s_{i,j}=1$ when $t_{i,j}=\{0,1\}$. On the other hand a relational similarity within a class can be defined as $r_{i,j}=0$ when $t_{i,j}=1$ and $r_{i,j}=1$ when $t_{i,j}=0$. $r_{i,j}$ is not defined for $t_{i,j}=2$. The complete approach is presented in Fig.~\ref{fig3a} and described subsequently.

\subsection{Architecture of feature representation learning and associated networks}

A CNN represented as $\mathtt{net_f(\cdot)}$ is employed to learn feature representation in an image. We employ a network similar to as used in~\cite{Cao_2018_CVPR} which is a modified version of AlexNet~\cite{Krizhevsky:2012:ICD:2999134.2999257}. The first 7 learnable layers are preserved and the output obtained then is represented as $\mathbf{z}$. This is fed subsequently to a classifier $\mathtt{net_C(\cdot)}$ which predicts the class of the clothing item as $\hat{\mathbf{y}}$ which is a one-hot vector. $\mathtt{net_C(\cdot)}$ consists of 3 fully connected layer arranged as $256-128-N$ where $N$ denotes the number of classes of clothes being looked into. The tensor $\mathbf{z}$ is also fed through a fully-connected layer $\mathtt{fcH(\cdot)}$ for generating the $K$-element long hashing tensor which is subsequently passed through a $\mathtt{tanh(\cdot)}$ function to generate the binary hash code $\mathbf{h}_i$ corresponding to an image $\mathbf{X}_i$. The discriminator network $\mathtt{net_D(\cdot)}$ consists of 1 convolutional layer with $1\times 1$ kernels followed by 4 fully connected layers  arranged as $128-256-128-1$ with sigmoid activation function used in the last layer.

\subsection{Learning of the semantic hashing network}

The approach for learning this network consists of the following 3 stages executed in subsequence per epoch.

\textbf{Stage 1:} Given an image $\mathbf{X}_i$ and its corresponding class label $\mathbf{y}_i$ the objective is to minimize the classification loss $J_C(\cdot)$ with respect to the prediction $\hat{\mathbf{y}}_i$ obtained from $\mathtt{net_C(\cdot)}$, thereby updating parameters in $\mathtt{net_f(\cdot)}$ and $\mathtt{net_C(\cdot)}$ as illustrated in Fig.~\ref{fig3b}. This stage assists $\mathtt{net_f(\cdot)}$ to learn features characteristic of representing different clothes. $J_C(\cdot)$ is evaluated using cross entropy (CE) loss between $\hat{\mathbf{y}}_i$ and ${\mathbf{y}}_i$.

\textbf{Stage 2:} Given a pair of images $\mathbf{X}_i$ and $\mathbf{X}_j$ and their corresponding type identifier $t_{i,j}$, the learnable parameters in $\mathtt{net_f(\cdot)}$ and $\mathtt{fcH(\cdot)}$ are updated to minimize the Cauchy losses $J_{\sigma_1}(\cdot)$ and $J_{\sigma_2}(\cdot)$ as illustrated in Fig.~\ref{fig3c}. The subjective similarity is predicted as $\hat{s}_{i,j}$ using Cauchy probability function~\cite{Cao_2018_CVPR}

\begin{equation}
\begin{aligned}
\hat{s}_{i,j} = P(s_{i,j}=1|\mathbf{h}_i,\mathbf{h}_j)
\\
= \frac{\gamma}{\gamma + d(\mathbf{h}_i,\mathbf{h}_j)}
\end{aligned}
\end{equation}

\noindent where $\hat{s}_{i,j}$ is the predicted subjective similarity index, $\gamma$ is a scale parameter and $d(\mathbf{h}_i,\mathbf{h}_j)$ is the Hamming distance measure. Binary cross entropy (BCE) extended with the Cauchy probability function is used to calculate the loss and is termed as Cauchy cross entropy loss~\cite{Cao_2018_CVPR}. 

\begin{equation}
\begin{aligned}
J_{\sigma}(s_{i,j}, \hat{s}_{i,j}) \\
= -\sum_{s_{i,j} \in S} \bigg( s_{i,j} \log(\hat{s}_{i,j})  + (1-s_{i,j})\log(1 - \hat{s}_{i,j}) \bigg) \\
= \sum_{s_{i,j} \in S} \bigg( s_{i,j} \log \frac{d(\mathbf{h}_i,\mathbf{h}_j)}{\gamma}  + \log \bigg(1+\frac{\gamma}{d(\mathbf{h}_i,\mathbf{h}_j)}  \bigg) \bigg)
\end{aligned}
\end{equation}

\noindent where $J_{\sigma}$ is the Cauchy cross entropy loss, $\gamma$ is a hyper parameter, and the normalized hamming distance between two code vectors $\mathbf{h}_i$ and $\mathbf{h}_j$ is defined as

\begin{equation}
\begin{aligned}
d(\mathbf{h}_i,\mathbf{h}_j) = \frac{K}{4} \bigg|\bigg|{\frac{\mathbf{h}_i}{||{\mathbf{h}_i}||}-\frac{\mathbf{h}_j}{||{\mathbf{h}_j}||}}\bigg|\bigg|_2^2
\\
= \frac{K}{2}(1-\cos{(\mathbf{h}_i,\mathbf{h}_j)})
\end{aligned}
\end{equation}

\noindent where $K$ denotes the bit length of the binary hash code. The loss $J_{\sigma_1}(\cdot)$ is minimized to obtain best \emph{subjective similarity} for all possible image pairs with $t_{i,j}\in\{0,1,2\}$. While minimizing \emph{relational similarity}, $J_{\sigma_2}(\cdot)$ is minimized for image pairs with $t_{i,j}\in\{0,1\}$ and not assessed for $t_{i,j}=2$. Learnable parameters of only $\mathtt{net_f(\cdot)}$ and $\mathtt{fcH(\cdot)}$ are updated in the process with $\alpha_1$ and $\alpha_2$ being relative weights associated with $J_{\sigma_1}(\cdot)$ and $J_{\sigma_2}(\cdot)$ respectively.

The $tanh(\cdot)$ function is used during the training to generate binary hash codes. However, it is not used during inference and is replaced directly with a sign based binarizer. 

\textbf{Stage 3:} Following Fig.~\ref{fig3d}, the hash codes $\mathbf{h}_i$ and $\mathbf{h}_j$ that are generated corresponding to an input image pair $\mathbf{X}_i$ and $\mathbf{X}_j$, are concatenated with channel shuffling in place. Given $\{i,j\}$ as the channel ordering at input to the shuffler, when shuffling takes places the channel ordering in output is $\{j,i\}$, else it remains same as $\{i,j\}$. The task of $\mathtt{net_D(\cdot)}$ is to identify if the shuffler had performed a shuffling operation and learning of parameters in $\mathtt{net_D(\cdot)}$ minimizes $J_D(\cdot)$. Since this stage is invoked only when $t_{i,j}=0$, and the objective being to have $\mathbf{h}_i$ and $\mathbf{h}_j$ as closest Hamming distance neighbours, learnable parameters in $\mathtt{net_f(\cdot)}$ and $\mathtt{fcH(\cdot)}$ are updated adversarially to maximally confuse $\mathtt{net_D(\cdot)}$ and increase $J_D(\cdot)$ which is evaluated with BCE. $\beta$ denotes the relative weight of adversarial update of $\mathtt{net_f(\cdot)}$ and $\mathtt{fcH(\cdot)}$.

\subsection{Retrieval as an inference problem}
\label{subsec:retrieval}

On completion of the training process, every image $\mathbf{X}_i$ in the gallery set is converted to a corresponding $K$-bit binary hash code $\mathbf{h}_i$ on being processed through $\mathtt{net_f(\cdot)}$, $\mathtt{fcH(\cdot)}$ and a binarizer. Given a query image  $\mathbf{X}_j$, it is first converted to obtain a binary hash code $\mathbf{h}_j$. The normalized Hamming distance $d(\mathbf{h}_i,\mathbf{h}_j)$ is then calculated for the pair and the images $\{\mathbf{X}_i\}$ in gallery set are ranked in ascending order of $d(\mathbf{h}_i,\mathbf{h}_j)$. The images in the gallery set that have the least Hamming distance with the query image constitute the top retrievals as illustrated in Fig.~\ref{fig:hashing}. 

The performance of retrieval is evaluated based on the standard metric of mean average precision (mAP). Given the query set with images $\{\mathbf{X}_j\}$, the average precision (AP) $AP_j@p$ is calculated based on the top-$p$ retrievals, which correspond to the set of $p$ closest neighbours of $\mathbf{X}_j$ evaluated based on $d(\mathbf{h}_i, \mathbf{h}_j)$   

\begin{equation}
AP_j@p = \frac{\sum_{k=1}^{p}P_j(k) \delta_j(k)}{\sum_{{k}=1}^{p}\delta_j({k})}
\label{eqn:mAP}
\end{equation}

\noindent where $\delta(\cdot)$ is an indicator function holding values as $\delta_j(k)=1$ if the corresponding $k^{th}$ ranked retrieved image and query image pair has $t_{k,j}=0$ or $t_{k,j}=1$, otherwise $\delta_j(k)=0$. $P_j(k)$ is the precision value for top-$k$ retrieved images 
 
\begin{equation}
P_j(k) = \frac{\sum_{n=1}^{k}Rel(n)}{k}, 
\end{equation}

\noindent where $Rel(n)$ denotes the ground truth relevance between the query image $\mathbf{X}_j$ and the $n^{th}$ retrieved image $\mathbf{X}_n$ from the gallery upto $k$-closest neighbours. $Rel(n) = 1$ when $t_{n,j}\in\{0,1\}$ and $Rel(n) = 0$ otherwise.
The mean of $AP_j@p\forall j\in\{\mathbf{X}_j\}$ is represented as $mAP@p$ value of retrieval.

Mean AP for top most $p$ retrievals $(mAP@top-p)$ is calculated for a query $\mathbf{X}_j$ if at least one image in the top-$p$ retrieved results from the gallery belongs to the same class as the query. In that case $AP_{j}(p) = 1$  when $t_{p,j}\in\{0,1\}$ and $AP_j(p)=0$ otherwise.  $mAP@top-p$ is calculated as the mean over all possible $AP_j(p)\forall j\in \{\mathbf{X}_j\}$.

\section{Experiments}
\label{sec:expt}

\begin{figure}[t]
\centering
    \begin{subfigure}[Distribution of images in men's inventory.]{\includegraphics[width=0.8\linewidth]{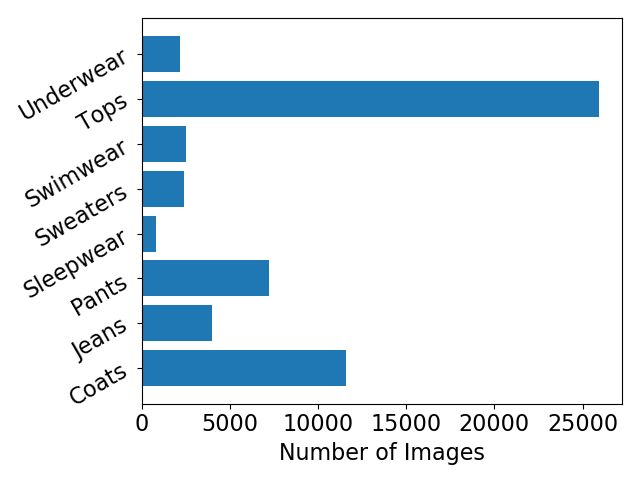}}
    \end{subfigure}
    \begin{subfigure}[Distribution of images in women's inventory.]{\includegraphics[width=0.8\linewidth]{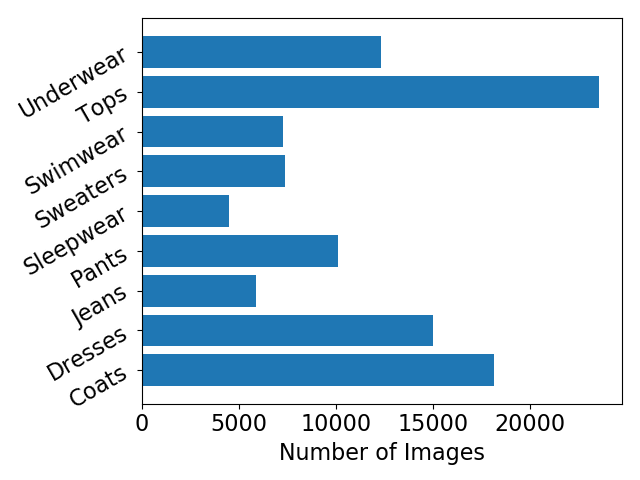}}
    \end{subfigure}
\caption{Distribution of various classes of clothing items in men's and women's inventory in the MVC dataset.}
\label{fig:cwdist}
\end{figure}

\subsection{Dataset}

\begin{figure}[t]
\begin{center}
\includegraphics[width=1.0\linewidth]{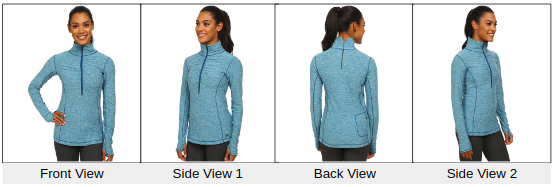}
\end{center}
   \caption{An example of images of the same clothing item under different pose variations. During training, any pair of images taken from this set would have $t_{i,j}=0$. During validation of retrieval performance, if any one of the images here constitutes a part of the Query set, then the remaining are part of the Gallery set.}
\label{fig:datasample}
\end{figure}

The performance of our scheme is experimentally validated using the \emph{MVC Dataset}~\cite{Liu2016MVCAD}, that is popularly used for benchmarking performance of view-invariant clothing item retrieval and clothing attribute prediction. The version of dataset used here consists of $161,260$ images each of size $1,920\times 2,240$ px. The dataset is provided as two subsets for Men and Women clothing items. The images are further manually filtered to remove wrong labelling and corrupted files,  to obtain $56,604$ images of men's clothing items and $104,010$ images of women's clothing item. Men's clothing items constitute of 8 classes viz. coats, pants, jeans, sleep wear, sweaters, swim wear, shirts tops, and underwear. Women's clothing items constitute of 9 classes viz. coats, jeans, pants, dresses, sleep wear, sweaters, swimwear, underwear, and tops. The distribution of these items is detailed in Fig.~\ref{fig:cwdist}. The images are distributed into Test, Train, Gallery and Query sets. Train set comprises of $60\%$ of total dataset, Test set comprises of $20\%$ of the elements. These together are used during the training process. The performance validation is performed on a Query and Gallery set where alternate poses of a clothing item present in the Query set make up the Gallery set, but there are no common images between these sets, and all the 4 sets are non-intersection sets, as illustrated with Fig.~\ref{fig:datasample}.

The training was carried out on men's and women's clothing items separately, and both combined together. \textbf{Men's clothing item} experiments are performed where the network is trained using randomly selected images from different classes paired with other randomly selected images. A total of $200,000$ combinations of paired images belonging to Type 2, $100,000$ pairs of Type 1 and $28,000$ pairs of Type 0 created from the training dataset. The loss functions are defined to be able to handle this kind of a data imbalance.  \textbf{Women's clothing item} experiments are performed using a total of $200,000$ combinations of paired images belonging to Type 2, $100,000$ pairs of Type 1 and $48,500$ pairs of Type 0 created from the training dataset. \textbf{Combined clothing items} experiments are performed using a total of $200,000$ combinations of paired images belonging to Type 2, $100,000$ pairs of Type 1 and $76,800$ pairs of Type 0 created from the training dataset.

\begin{figure}[t!]
\begin{center}
        \includegraphics[width= \columnwidth]{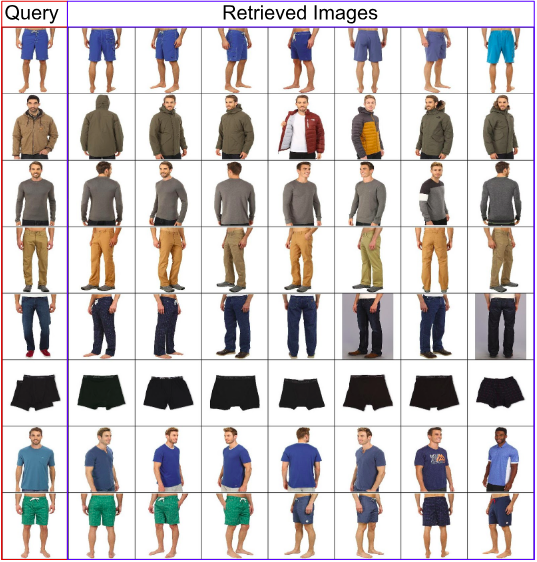}
        \caption{Men inventory retrieval result}
        \label{fig:shorta}
\end{center}
\end{figure}

\begin{figure}[t!]
\begin{center}
        \includegraphics[width= \columnwidth]{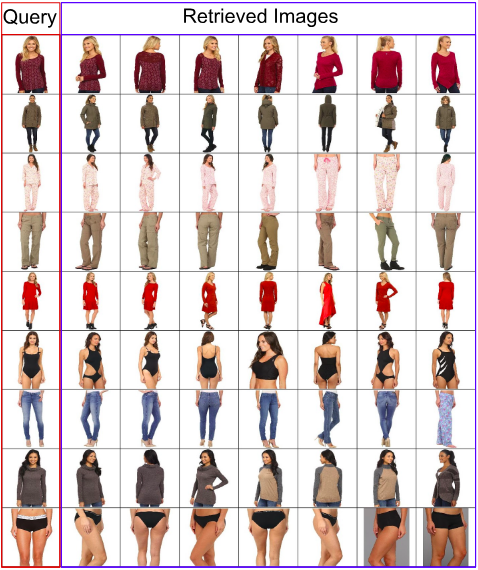}
        \caption{Women Inventory retrieval result}
        \label{fig:shortb}
\end{center}
\end{figure}

\begin{table*}[t!]
\caption{Performance Evaluation of the retrieval task for Men clothing inventory. \\[0.5ex]}
\centering
\begin{tabular}{m{2cm} m{2cm} m{2cm} m{2cm} m{2cm} m{2.1cm} m{2.1cm}|}
\hline
\textbf{Model} & \textbf{mAP@10} & \textbf{mAP@top-1} & \textbf{mAP@top-3} & \textbf{mAP@top-5} & \textbf{mAP@top-15 ($\ge 3$ hits)} & \textbf{mAP@top-15 ($\ge 5$ hits)}\\
\hline
\textbf{DMC-CD} &\textbf{90.65} & \textbf{95.20} & \textbf{98.17} & \textbf{98.63} & \textbf{97.94} & \textbf{86.98} \\
\textbf{DMC-C} &90.11 & 93.97 & 97.53 & 98.08 & 94.52 & 84.38 \\ \textbf{DMC}&84.13 &87.44 &96.11 &98.6 &93.83 &75.34\\\textbf{Vanilla \cite{Cao_2018_CVPR}}&53.26 &42.46 &68.9 &81.85 &65.29 &33.49\\
\hline
\end{tabular}
\label{tab:res1}
\end{table*}

\begin{table*}[t!]
\caption{Performance Evaluation of the retrieval task for Women Clothing Inventory. \\[0.5ex]}
\centering
\begin{tabular}{m{2cm} m{2cm} m{2cm} m{2cm} m{2cm} m{2.1cm} m{2.1cm}}
\hline
\textbf{Model} & \textbf{mAP@10} & \textbf{mAP@top-1} & \textbf{mAP@top-3} & \textbf{mAP@top-5} & \textbf{mAP@top-15 ($\ge 3$ hits)} & \textbf{mAP@top-15 ($\ge 5$ hits)}\\
\hline
\textbf{DMC-CD} &\textbf{82.67} &\textbf{85.55} &\textbf{96.11} &97.22 &91.11 &\textbf{74.44} \\
\textbf{DMC-C} &82.04 &85.05 &95.27 & \textbf{97.5} &\textbf{93.16} & 68.5 \\\textbf{DMC} &80.44 &84.16 &95.55 &97.44 &90.66 &61.18 \\
\textbf{Vanilla\cite{Cao_2018_CVPR}} &30.48 &19.04 &41.42 &56.67 &25.55 &5.77 \\
\hline
\end{tabular}
\label{tab:res2}
\end{table*}

\begin{table*}[t!]
\caption{Performance Evaluation of the retrieval task for MVC dataset. \\[0.5ex]}
\centering
\begin{tabular}{m{2cm} m{2cm} m{2cm} m{2cm} m{2cm} m{2.1cm} m{2.1cm}}
\hline
\textbf{Model} & \textbf{mAP@10} & \textbf{mAP@top-1} & \textbf{mAP@top-3} & \textbf{mAP@top-5} & \textbf{mAP@top-15 ($\ge 3$ hits)} & \textbf{mAP@top-15 ($\ge 5$ hits)}\\
\hline
\textbf{DMC-CD} &\textbf{83.88} &86.56 &\textbf{97.2} &\textbf{99.2} &95.65 &73.91 \\
\textbf{DMC-C} &83.73 &\textbf{88.14} &96.44 &98.44 &\textbf{96.04} & \textbf{75.09} \\
\textbf{DMC} &76.03 &78.46 &91.89 &96.34 &86.06 &54.0 \\
\textbf{Vanilla\cite{Cao_2018_CVPR}} &25.12 &13.04 &32.46 &47.56 &13.27 &1.01 \\
\hline
\end{tabular}
\label{tab:res3}
\end{table*}

\subsection{Training}

Pretrained weights of AlexNet~\cite{Krizhevsky:2012:ICD:2999134.2999257} used for solving the ImageNet for Large Scale Visual Recognition Challenge (ILSVRC)~\cite{ILSVRC15} task are used to initialize $\mathtt{net_f(\cdot)}$. $\mathtt{net_C(\cdot)}$, $\mathtt{fcH(\cdot)}$, and $\mathtt{net_D(\cdot)}$ were initialized with random weights. The images of size $1,920\times 2,240$ px were resized to $224 \times 224$ using bilinear interpolation to match the input size requirement for $\mathtt{net_f(\cdot)}$. The input images were horizontally flipped at random during training to induce view invariance in the learned model. Adam optimizer~\cite{DBLP:journals/corr/KingmaB14} was used during learning of parameters in $\mathtt{net_f(\cdot)}$, $\mathtt{net_C(\cdot)}$, $\mathtt{fcH(\cdot)}$, and $\mathtt{net_D(\cdot)}$ with learning rate of $10^{-5}$. The batch size was $256$ and the training continued till losses and accuracy trends across epochs were observed to saturate, at about ~$35$ epochs. The model parameters were defined as $\alpha_1=1$, $\alpha_2 = 1$, $\beta = 0.01$ and $\gamma = 3$. We had observed best performance for these parameters by varying $\gamma\in [2,50]$ following~\cite{Cao_2018_CVPR} and length of binary hash code is $K=48$. Experiments were performed on a Server with 2x Intel Xeon 4110 CPU, 12x8 GB DDR4 ECC Regd. RAM, 4 TB HDD, 4x Nvidia GTX 1080Ti GPU with 11 GB DDR5 RAM, and Ubuntu 16.04 LTS OS. The algorithms were implemented on Anaconda Python 3.7 with Pytorch 1.0.

\subsection{Results}

The experimental validation was performed separately for men's clothing items, women's clothing items and combined clothing items. Qualitative comparison of the performance in retrieving \textbf{men's clothing items} is presented in Fig.~\ref{fig:shorta} where each row corresponds to a class in the dataset and the first column in each row indicates a representative query image used, and subsequent 7 columns present the retrieved images. The results are quantitatively summarized in Table~\ref{tab:res1} as per measures detailed in Sec.~\ref{subsec:retrieval}.  In case of $mAP@top-15$, a successful hit is considered only if $(\geq 3)$ hits occur within the top $15$ retrieved results, and also if only $(\geq 5)$ hits occur. The different baselines considered include the following. \textbf{Vanilla}~\cite{Cao_2018_CVPR} is directly implemented as per prior art. Deep multi-stage Cauchy (\textbf{DMC}) is implemented with only $\mathtt{net_f(\cdot)}$ and $\mathtt{fcH(\cdot)}$ and learning to minimize only $J_{\sigma_1}(\cdot)$ and $J_{\sigma_2}(\cdot)$. \textbf{DMC-C} includes the classifier $\mathtt{net_C(\cdot)}$ along with the configuration of DMC and also looks to minimize $J_C(\cdot)$. \textbf{DMC-CD} includes the discriminator $\mathtt{net_D(\cdot)}$ along with DMC-C and while the optimizer on $\mathtt{net_D(\cdot)}$ works to minimize $J_D(\cdot)$, the optimization of $\mathtt{net_f(\cdot)}$ and $\mathtt{fcH(\cdot)}$ maximizes $J_D(\cdot)$ as an adversarial learning approach.

Similarly the qualitative performance in retrieving \textbf{women's clothing items} is presented in Fig.~\ref{fig:shortb} and quantitatively summarized in Table~\ref{tab:res2}. Similarly retrieval performance in \textbf{combined clothing items} is summarized in Table~\ref{tab:res3}. Across each of the sets of experiments it can be clearly observed that inclusion of a classifier, Cauchy cross entropy loss and finally a discriminator for adversarial learning has significantly improved the performance of retrieval by enabling generation of characteristic binary hash codes.

\begin{figure}[b!]
\centering
    \begin{subfigure}[Vanilla Cauchy Hashing]{
        \includegraphics[width=0.45\columnwidth]{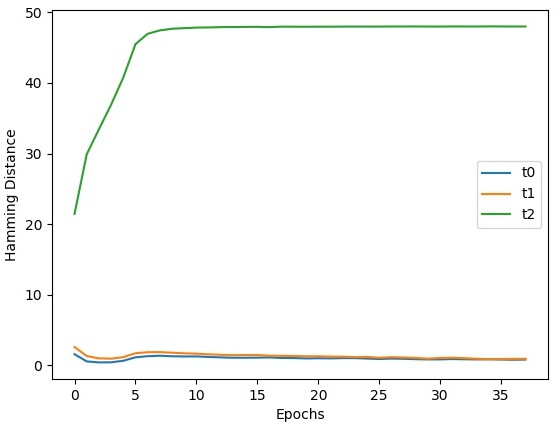}
        \label{fig:hdcmpvanilla}}
        \end{subfigure}
    \begin{subfigure}[DMC Hashing]{
    \includegraphics[width=0.45\columnwidth]{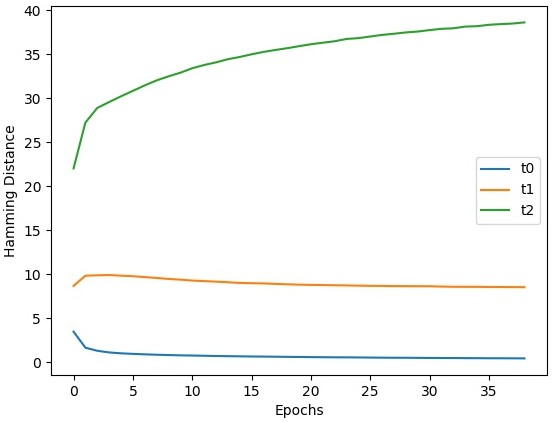}
        \label{fig:hdcmpdmc}}
    \end{subfigure}
\caption{Figure shows the relation between hamming distance and number of epochs of training performed.}
\label{fig:hdcmp}
\end{figure}

\begin{figure*}[t!]
\begin{center}
\centering
   \hfill
   \includegraphics[width=0.6\textwidth]{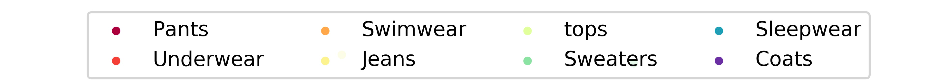}\hfill\break
   \begin{subfigure}[Vanilla(Men)]{
       \includegraphics[width=0.23\textwidth]{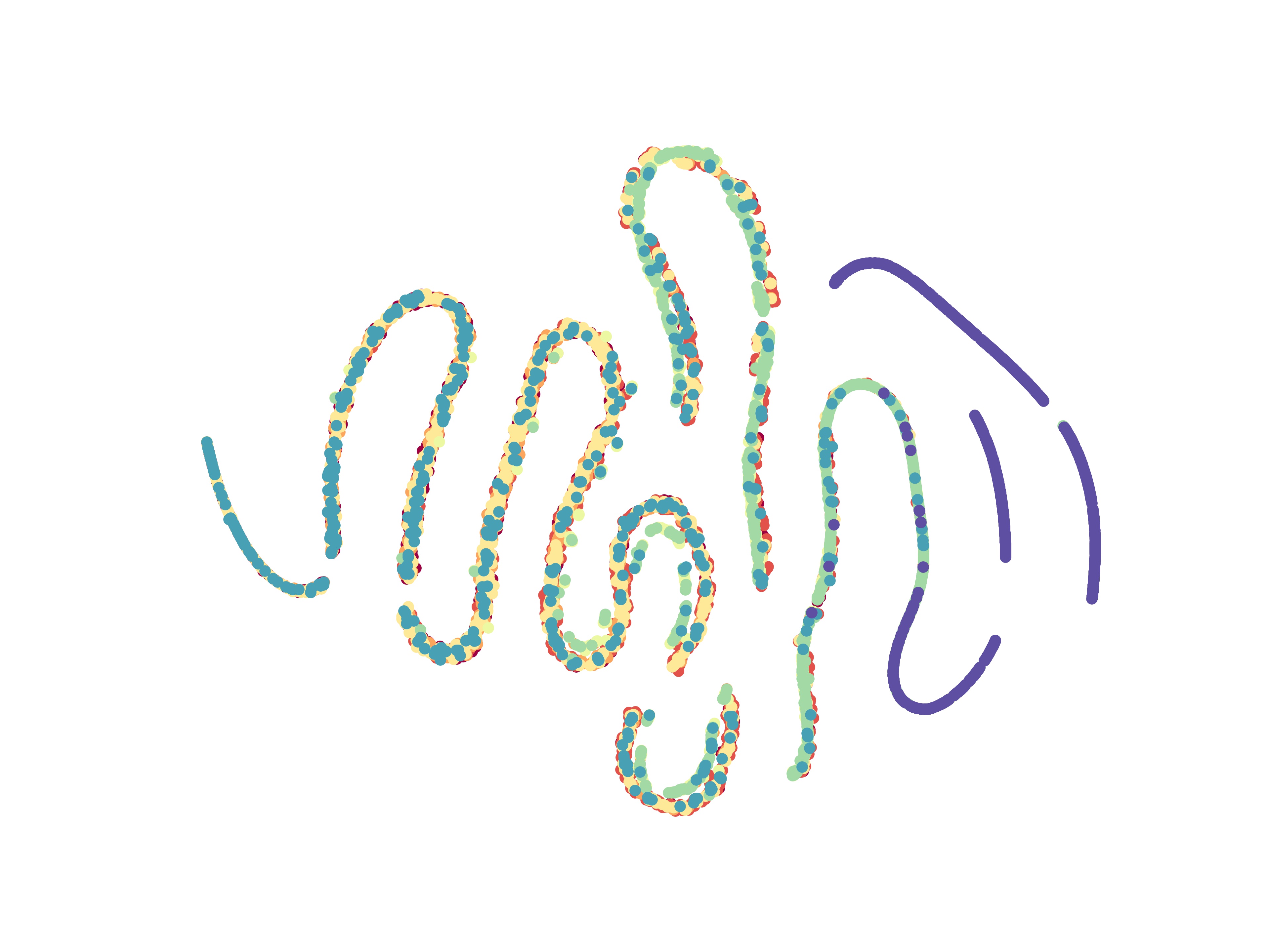}
       }
   \end{subfigure}
   \begin{subfigure}[DMC(Men)]{
       \includegraphics[width=0.23\textwidth]{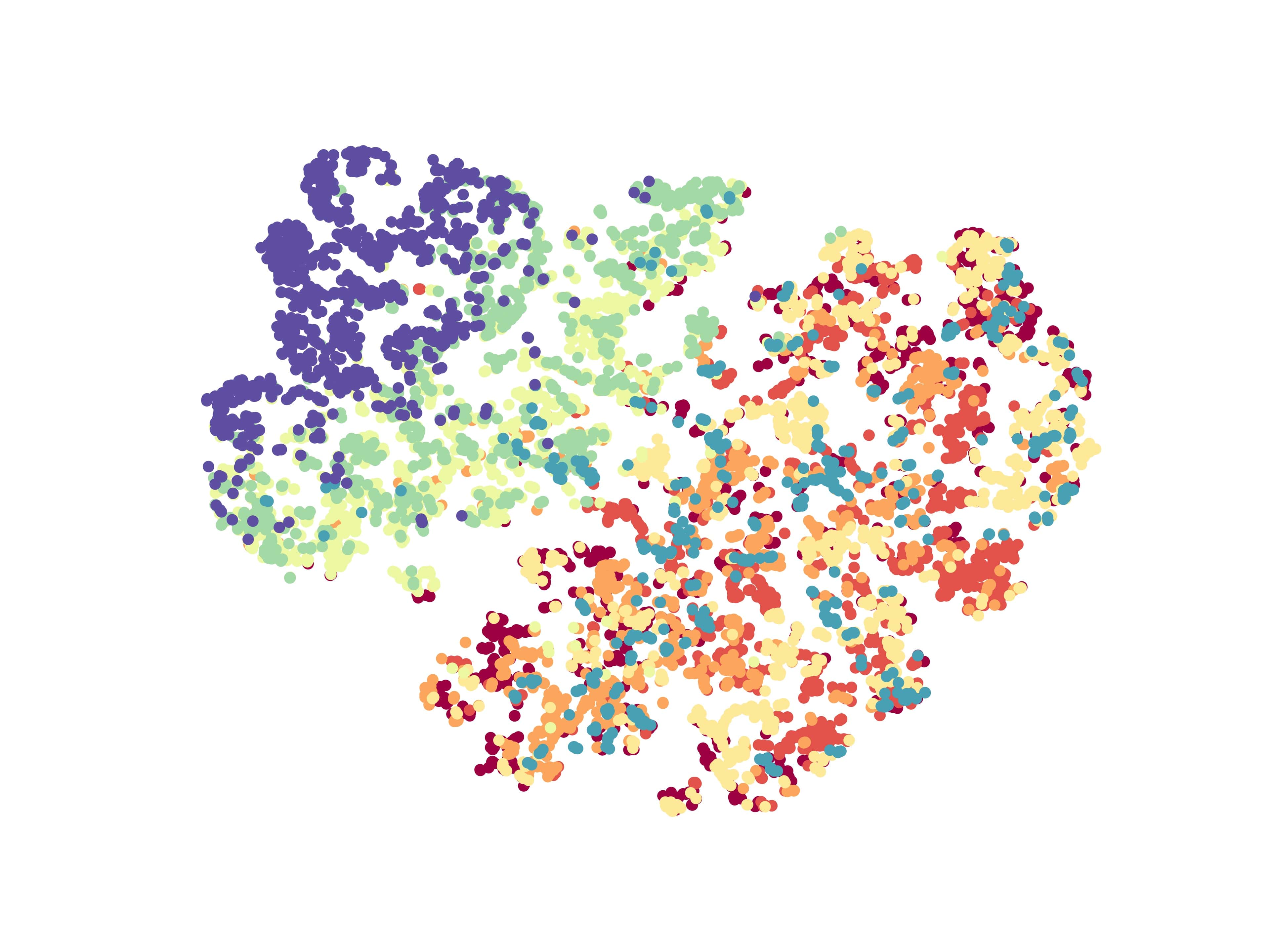}
       }
   \end{subfigure}
   \begin{subfigure}[DMC-C(Men)]{
       \includegraphics[width=0.23\textwidth]{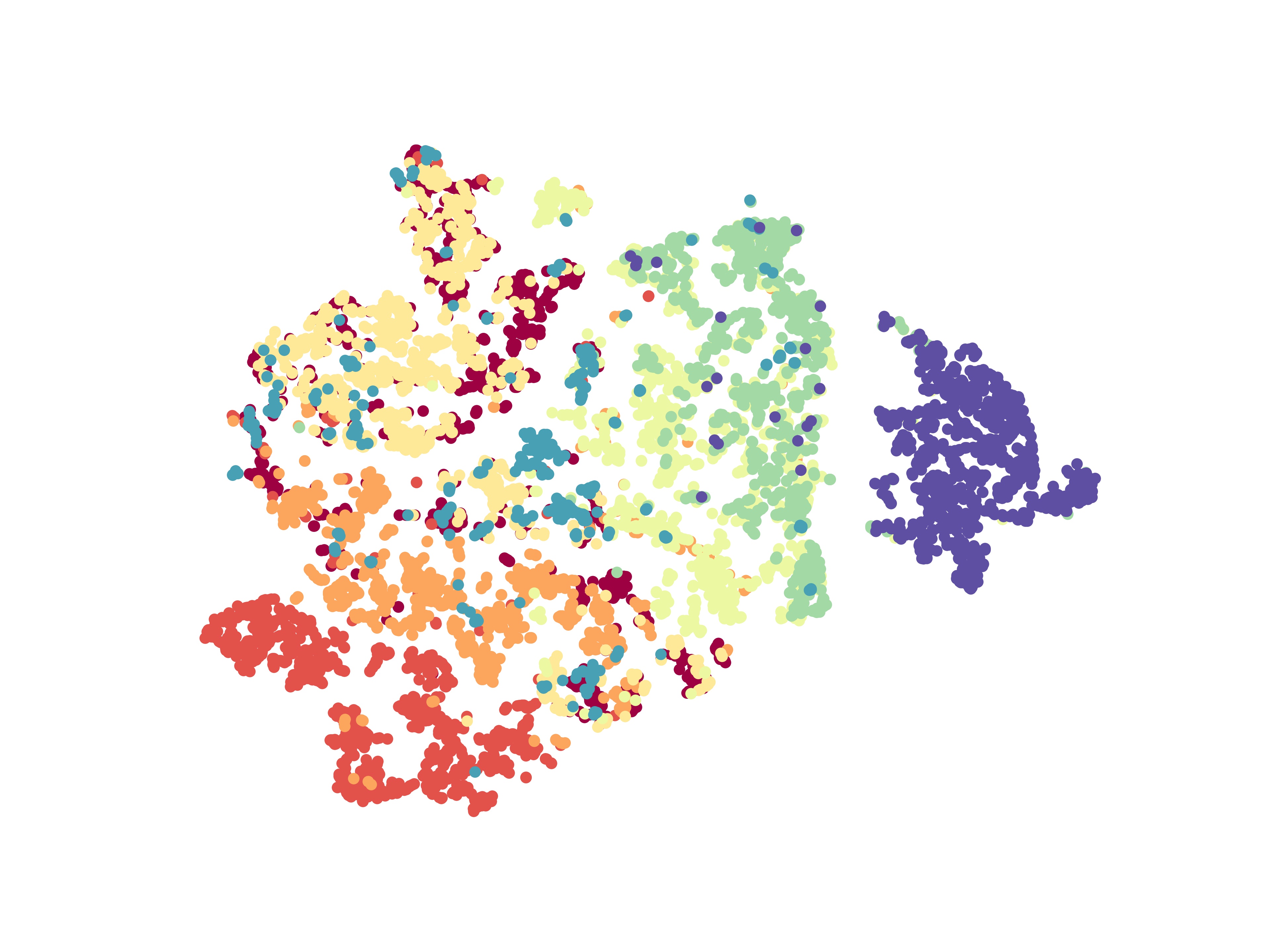}
       }
   \end{subfigure}
   \begin{subfigure}[DMC-CD(Men)]{
       \includegraphics[width=0.23\textwidth]{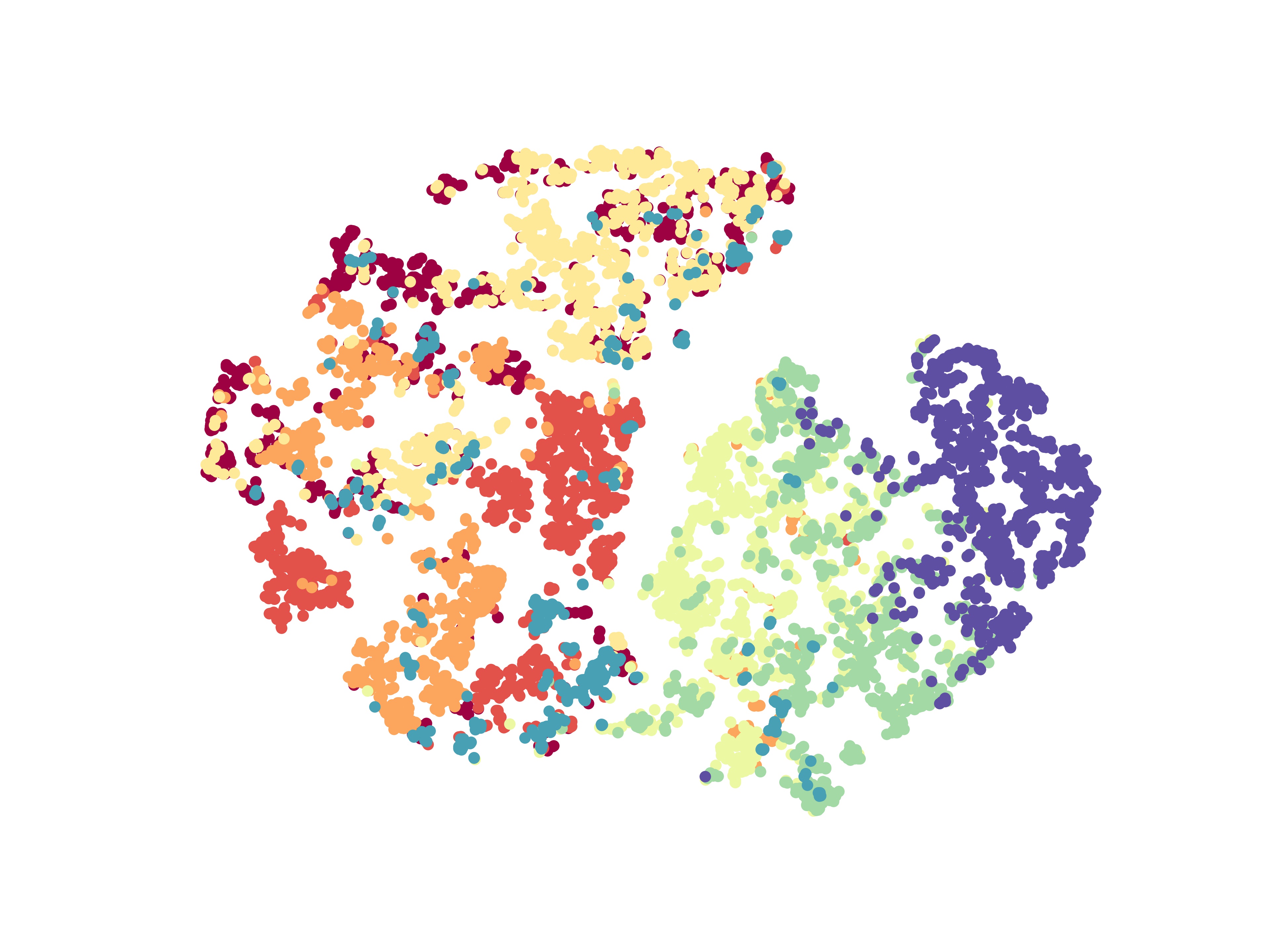}}
   \end{subfigure}

\centering
\hfill
\includegraphics[width=0.7\textwidth]{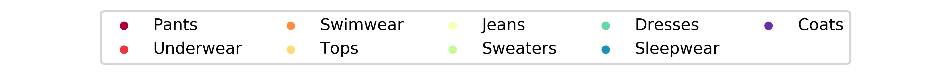}\hfill\break
   \begin{subfigure}[Vanilla(Women)]{
       \includegraphics[width=0.23\textwidth]{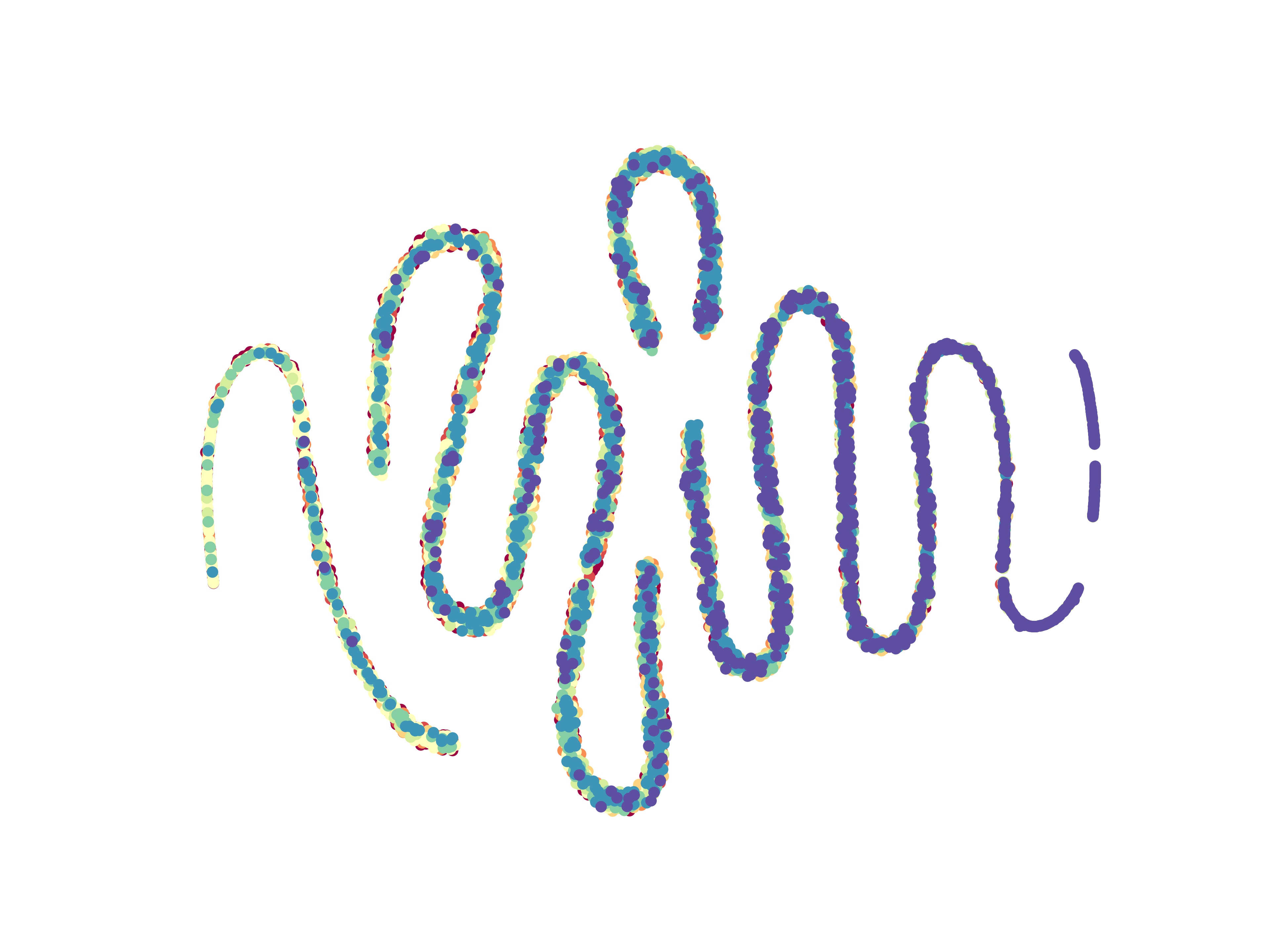}}
   \end{subfigure}
   \begin{subfigure}[DMC(Women)]{
       \includegraphics[width=0.23\textwidth]{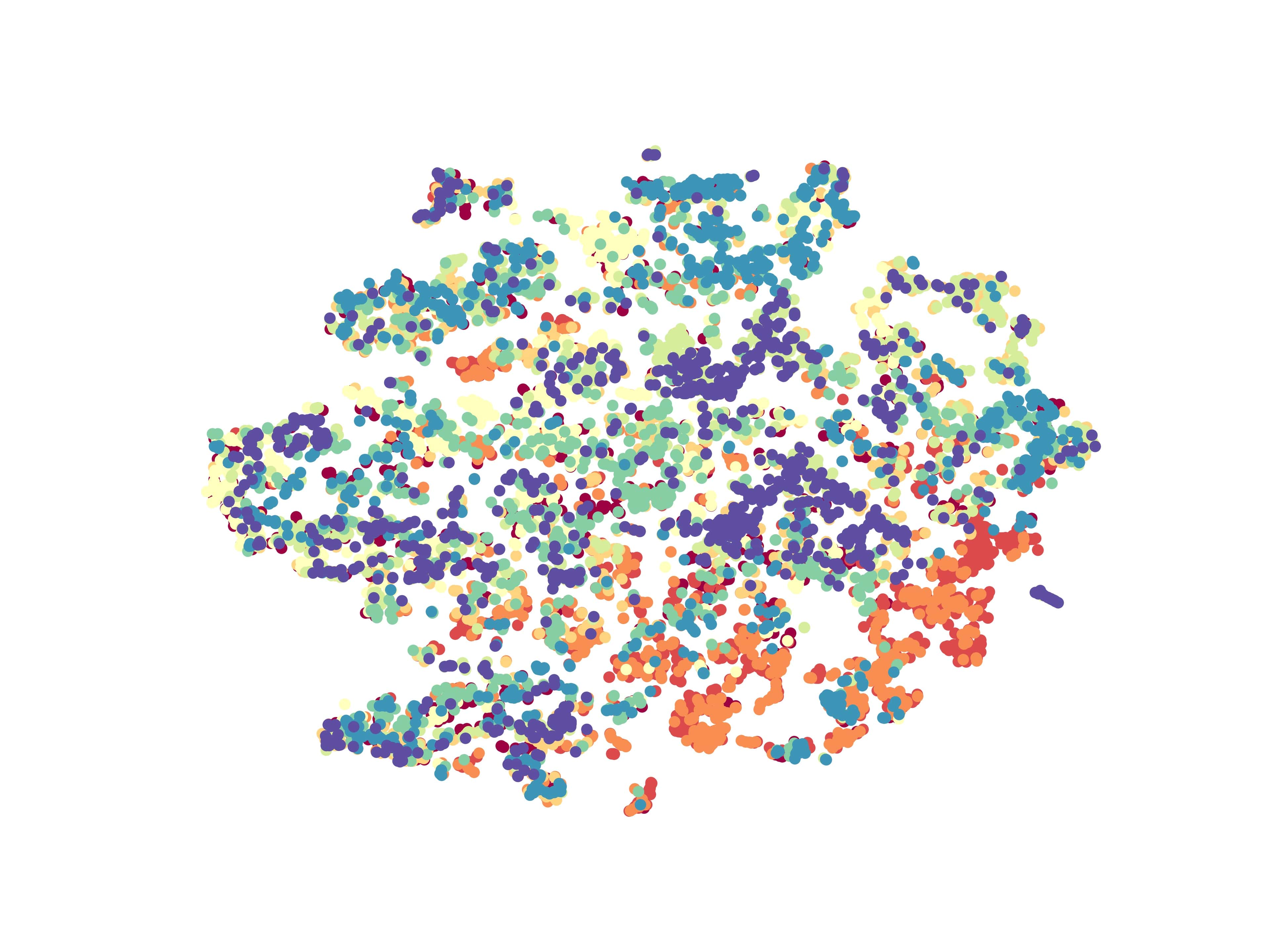}}
   \end{subfigure}
   \begin{subfigure}[DMC-C(Women)]{
       \includegraphics[width=0.23\textwidth]{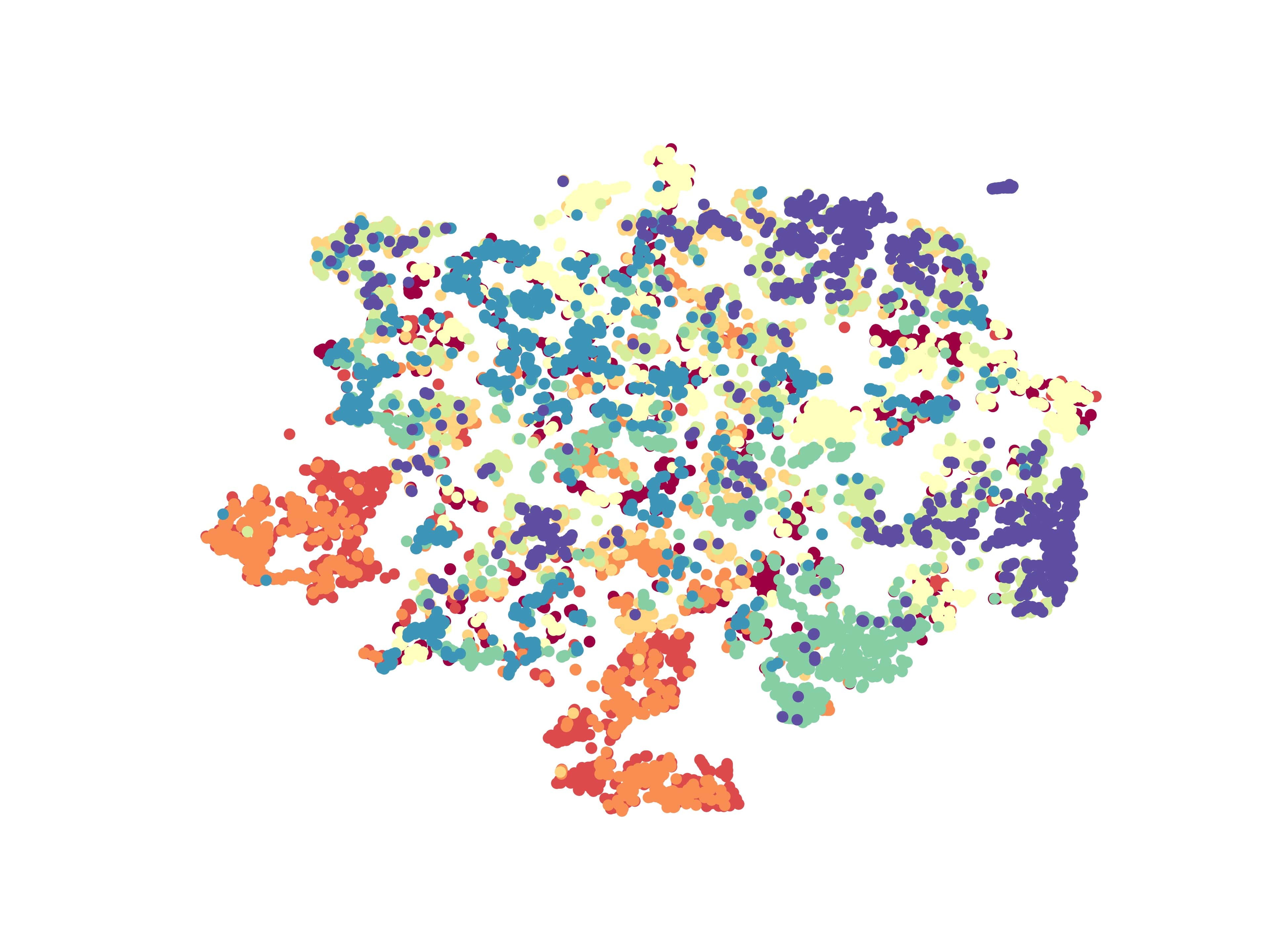}}
   \end{subfigure}
   \begin{subfigure}[DMC-CD(Women)]{
       \includegraphics[width=0.23\textwidth]{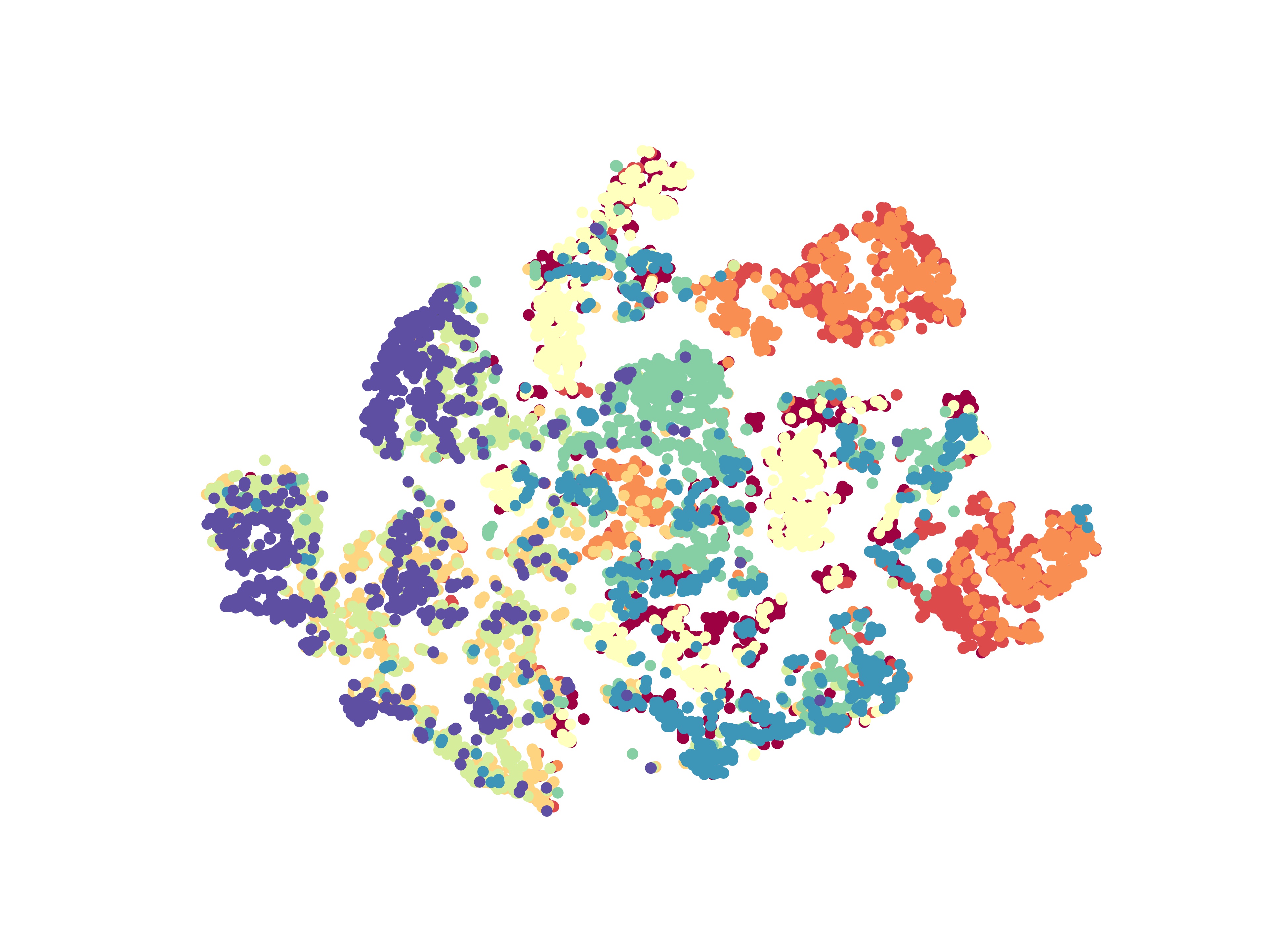}}
   \end{subfigure}

\centering
\hfill
\includegraphics[width=0.8\textwidth]{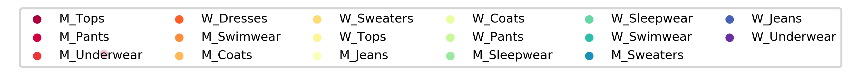}\hfill\break
   \begin{subfigure}[Vanilla(MVC)]{
       \includegraphics[width=0.23\textwidth]{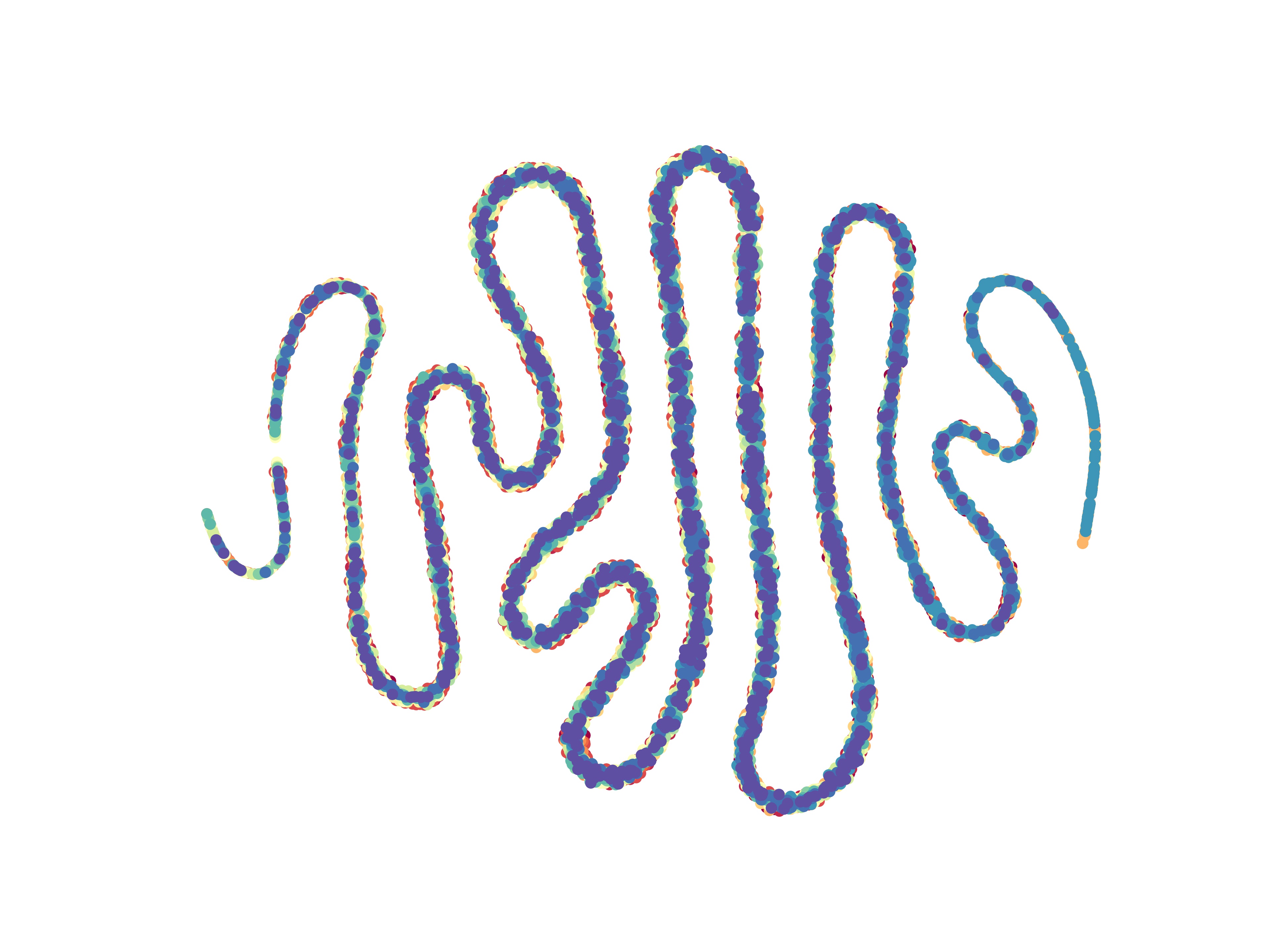}}
   \end{subfigure}
   \begin{subfigure}[DMC(MVC)]{
       \includegraphics[width=0.23\textwidth]{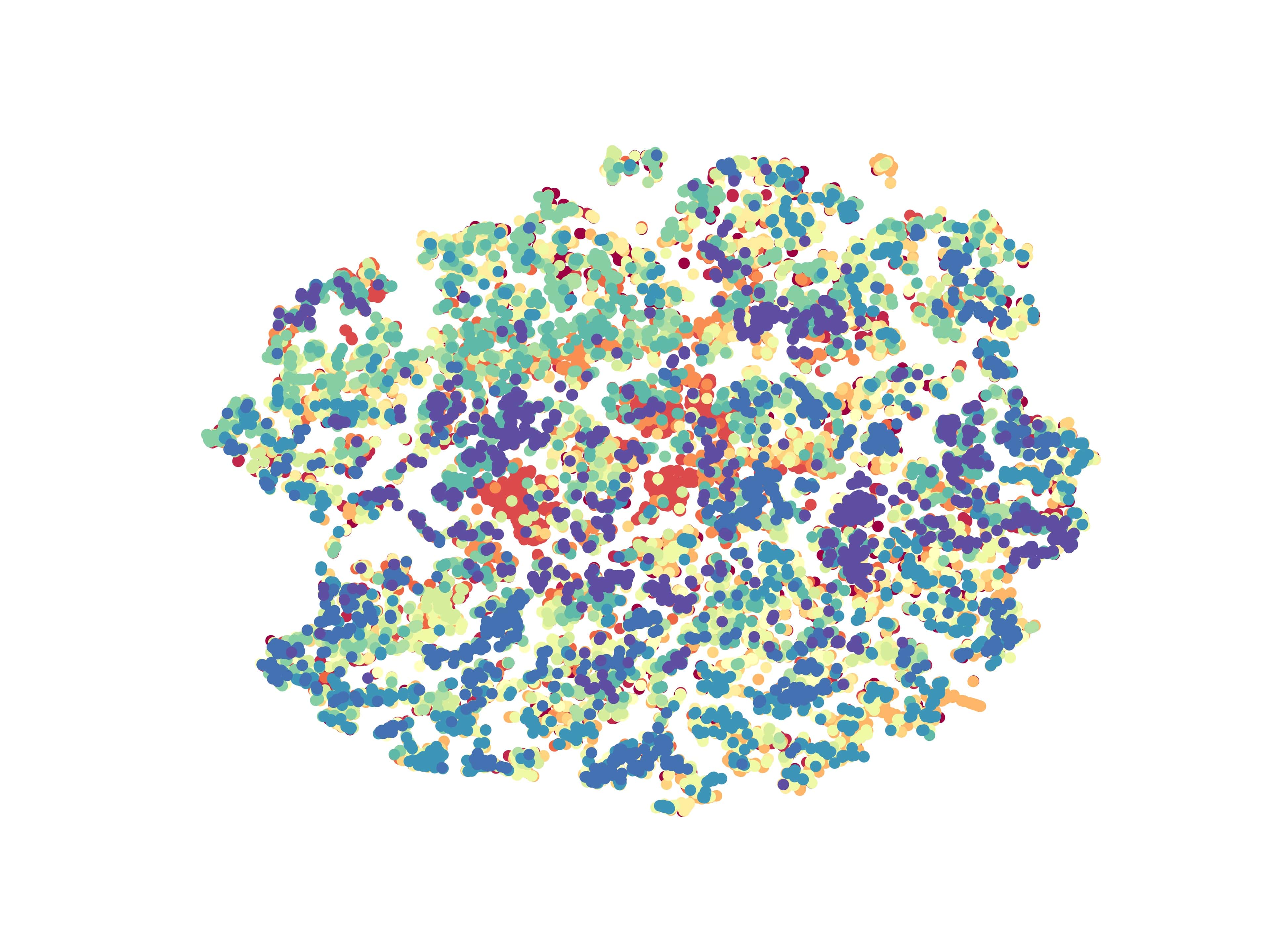}}
   \end{subfigure}
   \begin{subfigure}[DMC-C(MVC)]{
       \includegraphics[width=0.23\textwidth]{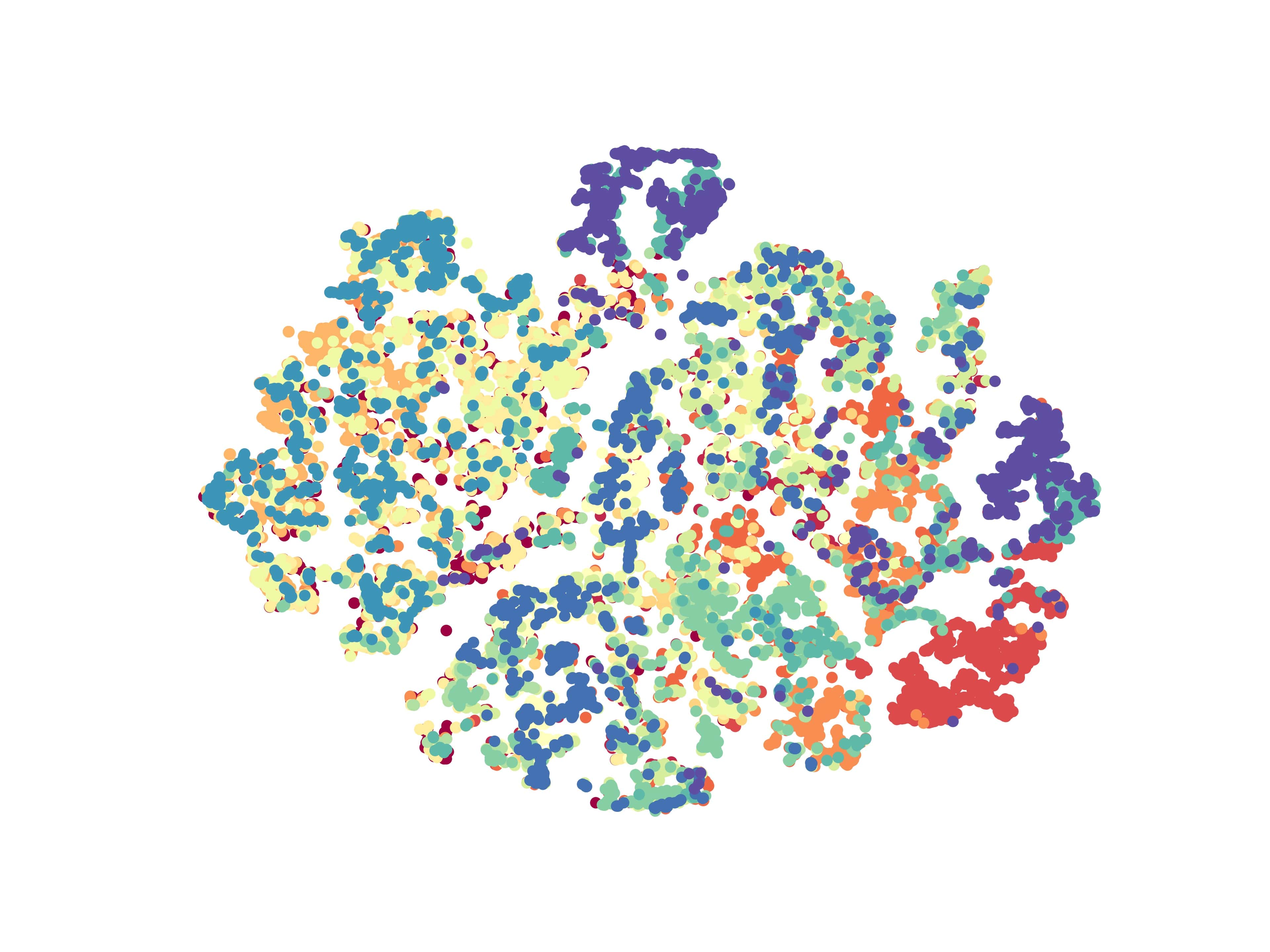}}
   \end{subfigure}
   \begin{subfigure}[DMC-CD(MVC)]{
       \includegraphics[width=0.23\textwidth]{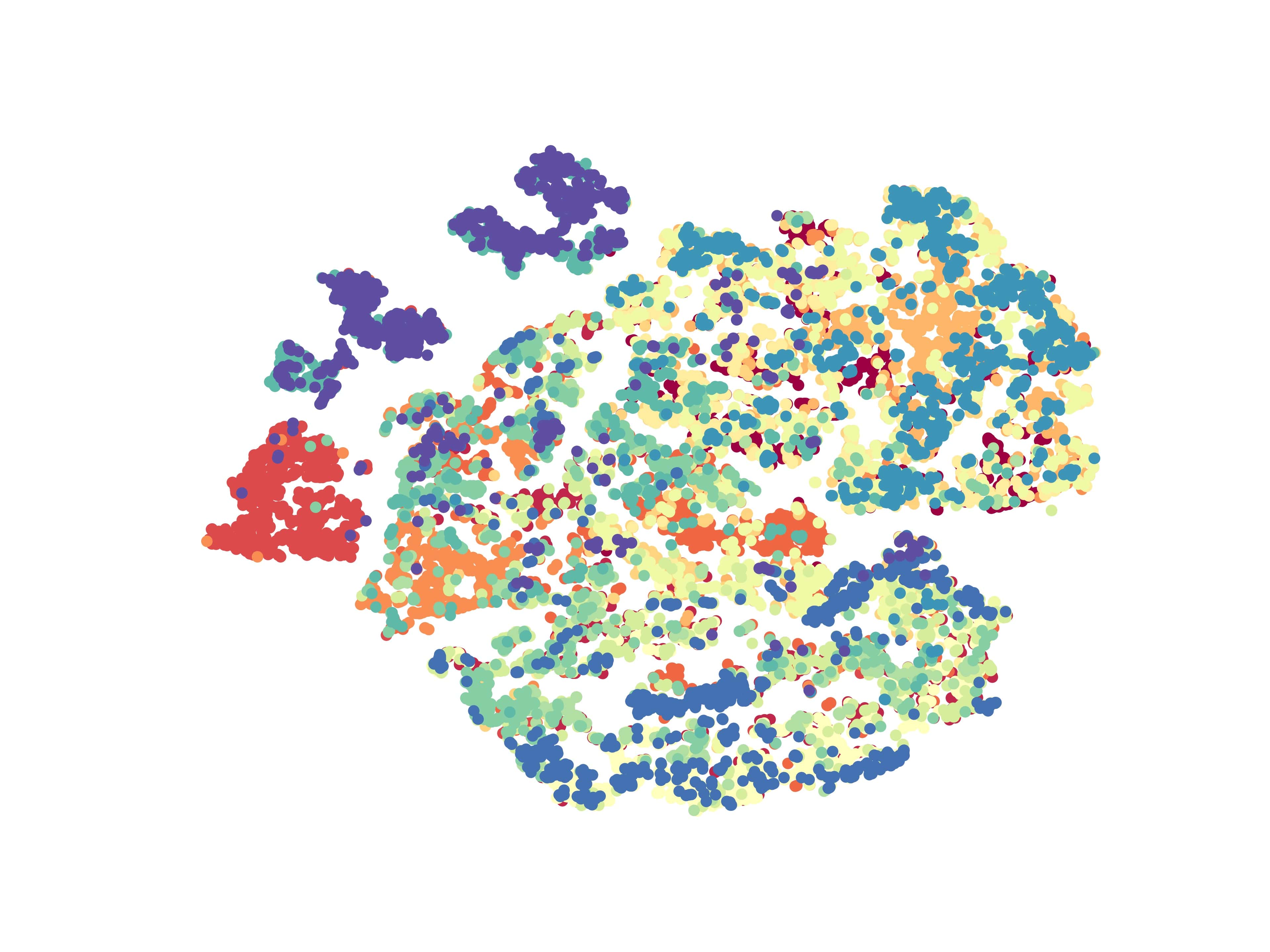}}
   \end{subfigure}
\end{center}
\caption{The t-SNE visualizations for the proposed architecture and its variants for hash codes generated using MVC dataset}
\label{fig:tsne}
\end{figure*}

\subsection{Discussion}
\label{sec:disc}

\subsubsection{\textbf{Learning with two Cauchy cross entropy losses}\newline}

As compared to learning with only $J_{\sigma_1}(\cdot)$ which is similar to the Vanilla~\cite{Cao_2018_CVPR} approach of increasing Hamming distance based separation margin between samples of Type 2, enabling learning by also including $J_{\sigma_2}(\cdot)$ increases the separation margin between the hash codes for samples of Type 1. This can be clearly observed in Fig.~\ref{fig:hdcmpvanilla} where across epochs of training, the separation between samples of Type 2 is very high by using only $J_{\sigma_1}(\cdot)$ but no significant difference is observed for samples of Type 1 from Type 0, which is possible with inclusion of $J_{\sigma_2}(\cdot)$ as can be observed in Fig.~\ref{fig:hdcmpdmc}. This  is possible due to the increase in spectral spread of $\mathbf{z}$ generated by $\mathtt{net_f(\cdot)}$ as can be observed in the tSNE plots in Fig.~\ref{fig:tsne}. Use of DMC forces increase in spectral spread, away from being focally concentrated around manifold distribution of $\mathbf{z}$ observed in the vanilla implementation.

\subsubsection{\textbf{Learning with a classifier}\newline}

The feature learning network $\mathtt{net_f(\cdot)}$ is generally initialized with weights from a network used to perform ImageNet classification task and is suited to represent natural image characteristics. While features obtained in $\mathbf{z}$ may not be characteristic to discriminate the different classes of images present, including $\mathtt{net_C(\cdot)}$ while optimizing its weights along with that of $\mathtt{net_f(\cdot)}$ while minimizing $J_C(\cdot)$ helps to obtain features characteristic of different classes of clothing items. This helps to improve performance by resulting in characteristic features for each class of clothing item and these features tend to exhibit clustering behaviour as seen with DMC-C in Fig.~\ref{fig:tsne}.

\subsubsection{\textbf{Adversarial learning with a discriminator}\newline}

One of the aspects desirable of the generated hash codes is that they are pose and view invariant for the same item. Essentially this implies that all images in Fig.~\ref{fig:datasample} should have the same hash code. We have achieved this by using the discriminator $\mathtt{net_D(\cdot)}$ with the purpose to identify if the first channel corresponds to $\mathbf{h}_i$ and second corresponds to $\mathbf{h}_j$ or vice-versa. The purpose in adversarial learning is to optimize weights in $\mathtt{net_f(\cdot)}$ and $\mathtt{fcH(\cdot)}$ such that it leads to maximize confusion for $\mathtt{net_D(\cdot)}$ leading to increase in $J_D(\cdot)$. This leads to assigning of similar hash binary codes and $\mathbf{z}$ for items of Type 0. The tSNE plot in Fig.~\ref{fig:tsne}. exhibits the close clustering achieved with DMC-CD.

\section{Conclusion}
\label{sec:conc}

This work presents a Deep Multi Cauchy Hashing framework and its variants to perform view invariant fast subjective search in fashion inventory with high accuracy. In this direction, the work establishes a comparison between baseline DMC model and its variants in Table \ref{tab:res1}, \ref{tab:res2} and \ref{tab:res3}. The proposed  scheme maximizes the hamming distance between semantically dissimilar images and minimizes the same between semantically similar images. The formation of discriminative clusters as shown in figure \ref{fig:tsne} justifies the claim. Extensive experiments show that the model can show state of art performance as can be seen in results obtained on MVC Dataset in figures \ref{fig:shorta} and \ref{fig:shortb}. With rapid expansion of e-commerce, the proposed technique can be essential in retrieval tasks not limited to just fashion industry.


\bibliographystyle{ACM-Reference-Format}
\bibliography{ms.bbl}
\end{document}